# Variable sigma Gaussian processes: An expectation propagation perspective

**Yuan (Alan) Qi**
CS & Statistics Departments, Purdue University
alanqi@cs.purdue.edu

**Ahmed H. Abdel-Gawad**
ECE Department, Purdue University
aabdelga@purdue.edu

Thomas P. Minka
Microsoft Research Cambridge, UK
minka@microsoft.com

## Abstract

Gaussian processes (GPs) provide a probabilistic nonparametric representation of functions in regression, classification, and other problems. Unfortunately, exact learning with GPs is intractable for large datasets. A variety of approximate GP methods have been proposed that essentially map the large dataset into a small set of basis points. The most advanced of these, the variable-sigma GP (VSGP) (Walder et al., 2008), allows each basis point to have its own length scale. However, VSGP was only derived for regression. We describe how VSGP can be applied to classification and other problems, by deriving it as an expectation propagation algorithm. In this view, sparse GP approximations correspond to a KL-projection of the true posterior onto a compact exponential family of GPs. VSGP constitutes one such family, and we show how to enlarge this family to get additional accuracy. In particular, we show that endowing each basis point with its own *full covariance matrix* provides a significant increase in approximation power.

## 1 Introduction

Gaussian processes (GP) are powerful nonparametric Bayesian approach to modelling unknown functions. As such, they can be directly used for classification and regression (Rasmussen & Williams, 2006), or embedded into a larger model such as factor analysis (Teh et al., 2005), relational learning (Chu et al., 2006), or reinforcement learning (Deisenroth et al., 2009). Unfortunately, the cost of GPs can be prohibitive for large datasets. Even for the regression case where the GP prediction formula is analytic, training the exact GP model with $N$ points demands an $O(N^3)$ cost for inverting the covariance matrix and predicting a new output requires $O(N^2)$ cost in addition to storing all of the training points.

Ideally, we would like a compact representation, much smaller than the number of training points, of the posterior distribution for the unknown function. This compact representation could be used to summarize training data for Bayesian learning, or it could be passed around as a message in order to do inference in a larger probabilistic model. One successful approach is to map the training data into a small set of basis points, then compute the exact posterior that results from those points. The basis points could literally be a subset of the training instances (Csato, 2002; Lawrence et al., 2002) or they could be artificial "pseudo-inputs" that represent the training set (Snelson & Ghahramani, 2006). Furthermore, the GP that is used to for inference on the basis points need not match the original GP. In particular, as shown by Walder et al. (2008), the GP applied to the basis points should have a longer length scale than the original (since the data is now sparser). Their "variable sigma Gaussian process" (VSGP) algorithm (Walder et al., 2008) allows a different length scale for each basis point.

For classification problems, a further approximation step is needed. The training set needs to not only be reduced but also transformed from classification data to regression data. In this paper, we show how both of these approximation steps can be accomplished in the same framework of expectation propagation (EP). The novel contributions of this paper include:

- Derive VSGP and other sparse GP algorithms as EP algorithms employing a particular choice of approximating family,
- Extend VSGP to use a full covariance matrix around each basis point,
- Show how to apply the VSGP approximation to classification and other likelihoods, and
- Evaluate the benefit of full covariance matrices on classification and regression problems (section 5).

## 2  Gaussian process regression and classification

We denote $N$ independent and identically distributed samples as $\mathcal{D} = \{(\mathbf{x}_1, y_1), \ldots, (\mathbf{x}_n, y_n)\}_N$, where $\mathbf{x}_i$ is a $d$ dimensional input and $y_i$ is a scalar output. We assume there is a latent function $f$ that we are modeling and the noisy realization of latent function $f$ at $\mathbf{x}_i$ is $y_i$.

A Gaussian process places a prior distribution over the latent function $f$. Its projection at the samples $\{\mathbf{x}_i\}$ defines a joint Gaussian distribution:

$$p(\mathbf{f}) = \mathcal{N}(\mathbf{f}|\mathbf{m}^0, K)$$

where $m_i^0 = m^0(\mathbf{x}_i)$ is the mean function and $K_{ij} = K(\mathbf{x}_i, \mathbf{x}_j)$ is the covariance function, which encodes the prior notation of smoothness. Normally the mean function is simply set to be zero and we follow this tradition in this paper. A typical kernel covariance function is the squared exponential, also know as Radial Basis Function (RBF),

$$k(\mathbf{x}, \mathbf{x}') = \exp(-\frac{||\mathbf{x}' - \mathbf{x}||^2}{2\sigma^2}), \tag{1}$$

where $\sigma$ is a hyperparameter.

For regression, we use a Gaussian likelihood function

$$p(y_i|f) = \mathcal{N}(y_i|f(\mathbf{x}_i), v_y) \tag{2}$$

where $v_y$ is the observation noise. For classification, the data likelihood has the form

$$p(y_i|f) = (1-\epsilon)\phi(f(\mathbf{x}_i)y_i) + \epsilon\phi(-f(\mathbf{x}_i)y_i) \tag{3}$$

where $\epsilon$ models the labeling error and $\phi(\cdot)$ is a nonlinear function, ie., a cumulative Gaussian distribution or a step function such that $\phi(f(x_i)) = 1$ if $f(x_i) \leq 0$ and $\phi(f(x_i)) = 0$ otherwise.

Given the Gaussian process prior over $f$ and the data likelihood, the posterior process is

$$p(f|\mathcal{D}, \mathbf{t}) \propto GP(f|0, K) \prod_{i=1}^{N} p(y_i|f) \tag{4}$$

Since the Gaussian process is grounded on the $N$ examples, they are called the basis points.

For the regression problem, the posterior process has an analytical form. But to make a prediction on a new sample, we need to invert a $N$ by $N$ matrix. If the training set is big, this matrix inversion will be too costly. For classification or other nonlinear problems, the computational cost is even higher since we do not have a analytical solution to the posterior process and the complexity of the process grows with the number of training samples.

## 3  Adaptive sparse Gaussian processes

In this section we present the novel sparse Gaussian process model coupled with an efficient inference algorithm.

### 3.1 Model

To save the computational and memory cost, we approximate the exact Gaussian process model (4) by a sparse model parameterized by $(\mathbf{u}, \mathbf{b}, \mathbf{c}, \boldsymbol{\Lambda})$:

$$q(f) \propto GP(f|0, K) \prod_{k=1}^{M} \mathcal{N}(u_k | \int f(\mathbf{x}) \mathcal{N}(\mathbf{x}|b_k, c_k) \mathrm{d}\mathbf{x}, \lambda_k^{-1}) \tag{5}$$

In this sparse model, we can change the number of basis points, $M$, to regulate the model complexity. Normally, we set $M \ll N$. Importantly, the basis point $b_k$ is blurred by the Gaussian distribution $\mathcal{N}(\mathbf{x}|b_k, c_k)$, which can represent how the data distributed *locally* around $b_k$. The parameter $u_k$ represents a virtual regression target for $b_k$, and $\lambda_k$ is its precision.

The distribution (5) is a Gaussian process, so it has a mean and covariance function. To determine these, first consider the following functional:

$$Z(m^0) = \int GP(f|m^0, K) \mathcal{N}(\mathbf{u}|\mathbf{g}_B(f), \boldsymbol{\Lambda}^{-1}) \mathrm{d}f \tag{6}$$

where $\mathbf{u} = (u_1, \ldots, u_M)$, the subscript $B$ denote the basis set $(b_1, \ldots, b_M)$, and

$$\mathbf{g}_B(f) = \Big[ \int f(\mathbf{x}) \mathcal{N}(\mathbf{x}|b_1, c_1) \mathrm{d}\mathbf{x}, \ldots, \int f(\mathbf{x}) \mathcal{N}(\mathbf{x}|b_M, c_M) \mathrm{d}\mathbf{x} \Big]^{\mathrm{T}} \tag{7}$$

$$\boldsymbol{\Lambda}^{-1} = \mathrm{diag}(\lambda_1^{-1}, \lambda_2^{-1}, \ldots, \lambda_M^{-1}) \tag{8}$$

Now $\mathbf{g}_B(f)$ follows a joint Gaussian distribution with the mean $\tilde{\mathbf{m}}_B^0$ and $\tilde{\mathbf{V}}_B^0$:

$$\tilde{m}_{Bj}^0 = \int \int f(\mathbf{x}) \mathcal{N}(\mathbf{x}|b_j, c_j) GP(f|m^0, K) \mathrm{d}\mathbf{x} \mathrm{d}f = \int m^0(\mathbf{x}) \mathcal{N}(\mathbf{x}|b_j, c_j) \mathrm{d}\mathbf{x} \tag{9}$$

$$\tilde{V}_{Bij}^0 = \int \int \mathcal{N}(\mathbf{x}|b_i, c_i) K(\mathbf{x}, \mathbf{x}') \mathcal{N}(\mathbf{x}'|b_j, c_j) \mathrm{d}\mathbf{x} \mathrm{d}\mathbf{x}' \tag{10}$$

Therefore, we can compute $Z(m^0)$ as follows:

$$Z(m^0) = \int \mathcal{N}(\mathbf{u}|\mathbf{g}_B, \boldsymbol{\Lambda}^{-1}) \mathcal{N}(\mathbf{g}_B|\tilde{\mathbf{m}}_B^0, \tilde{\mathbf{V}}_B^0) \mathrm{d}f = \mathcal{N}(\mathbf{u}|\tilde{\mathbf{m}}_B^0, \beta^{-1}) \tag{11}$$

where $\beta = (\tilde{\mathbf{V}}_B^0 + \boldsymbol{\Lambda}^{-1})^{-1}$. Define $\alpha = \beta \mathbf{u}$. Then the mean and the covariance functions of the sparse model are characterized by $\alpha$ and $\beta$. In particular, we have the following theorem to describe the relationship between them.

**Theorem 1** *The posterior process $q(f)$ defined in (5) has the mean function $m(\mathbf{x})$ and covariance function $V(\mathbf{x}, \mathbf{x}')$:*

$$m(\mathbf{x}) = \tilde{K}(\mathbf{x}, B) \alpha \tag{12}$$

$$V(\mathbf{x}, \mathbf{x}') = K(\mathbf{x}, \mathbf{x}') - \tilde{K}(\mathbf{x}, B) \beta \tilde{K}(B, \mathbf{x}') \tag{13}$$

where $\tilde{K}(\mathbf{x}, B) = [\tilde{K}(\mathbf{x}, b_1), \ldots, \tilde{K}(\mathbf{x}, b_j), \ldots, \tilde{K}(\mathbf{x}, b_M)]$, $\tilde{K}(B, \mathbf{x}) = (\tilde{K}(\mathbf{x}, B))^{\mathrm{T}}$, and $\tilde{K}(\mathbf{x}, b_j) = \int K(\mathbf{x}, \mathbf{x}') \mathcal{N}(\mathbf{x}'|b_j, c_j) \mathrm{d}\mathbf{x}'$.

When using the RBF covariance function (1), we have

$$\tilde{K}(x, B) = (2\pi\sigma^2)^{M/2} [\mathcal{N}(\mathbf{x}|b_1, c_1 + \sigma^2 \mathbf{I}), \ldots, \mathcal{N}(\mathbf{x}|b_M, c_M + \sigma^2 \mathbf{I})]. \tag{14}$$

**Proof:** First, consider the minimization of the KL divergence between the posterior process $q(f) \propto GP(f|m^0, K) \mathcal{N}(\mathbf{u}|\mathbf{g}_B(f), \boldsymbol{\Lambda}^{-1})$ for the sparse model and a process $\bar{q}(f)$ in the exponential family. Since $q(f)$ also belongs to the exponential family, this minimization will achieve the optimal solution, i.e., $\bar{q}(f) = q(f)$.

Now to obtain the mean function $m(\mathbf{x})$ and the covariance function $V(\mathbf{x}, \mathbf{x}')$ for $q(f)$, we can solve $\bar{q}(f)$ by the KL minimization. This leads to the following moment matching equations:

$$m(\mathbf{x}) = m^0(\mathbf{x}) + \int K(\mathbf{x}, \mathbf{a}') \frac{\mathrm{d} \log Z}{\mathrm{d} m^0(\mathbf{a}')} \mathrm{d}\mathbf{a}' \tag{15}$$

$$V(\mathbf{x}, \mathbf{x}') = K(\mathbf{x}, \mathbf{x}') + \int\int K(\mathbf{x}, \mathbf{a}) \frac{\mathrm{d}^2 \log Z}{\mathrm{d} m^0(\mathbf{a}) \mathrm{d} m^0(\mathbf{a}')} K(\mathbf{a}', \mathbf{x}') \mathrm{d}\mathbf{a}\mathrm{d}\mathbf{a}' \tag{16}$$

Based on (11), it is easy to obtain

$$\frac{\mathrm{d} \log Z}{\mathrm{d} m^0(\mathbf{a})} = \frac{\mathrm{d}}{\mathrm{d} \tilde{\mathbf{m}}_B} \big( -\frac{1}{2}(\mathbf{u} - \tilde{\mathbf{m}}_B)^{\mathrm{T}} \beta (\mathbf{u} - \tilde{\mathbf{m}}_B) \big) \frac{\mathrm{d} \tilde{\mathbf{m}}_B}{\mathrm{d} m^0(\mathbf{a})} \tag{17}$$

$$= [\mathcal{N}(\mathbf{x}|b_1, c_1), \ldots, \mathcal{N}(\mathbf{x}|b_M, c_M)] \beta (\mathbf{u} - \tilde{\mathbf{m}}_B) \tag{18}$$

Combining (15) and (18) gives

$$m(\mathbf{x}) = m^0(\mathbf{x}) + \int K(\mathbf{x}, \mathbf{x})[\mathcal{N}(\mathbf{x}|b_1, c_1), \ldots, \mathcal{N}(\mathbf{x}|b_M, c_M)] \beta (\mathbf{u} - \boldsymbol{\rho}) \mathrm{d}\mathbf{x} \tag{19}$$

$$= m^0(\mathbf{x}) + \tilde{K}(\mathbf{x}, B) \beta (\mathbf{u} - \boldsymbol{\rho}) = m^0(\mathbf{x}) + \tilde{K}(\mathbf{x}, B)(\alpha - \beta \tilde{\mathbf{m}}_B^0) \tag{20}$$

where $\tilde{K}(\mathbf{x}, B)$ is defined in (14). Setting the prior mean function $m^0(\mathbf{x}) = 0$, we have $\tilde{\mathbf{m}}_B^0 = 0$. As a result, equation (12) holds.

From (18) it follows that

$$\frac{\mathrm{d}^2 \log Z}{\mathrm{d} m^0(\mathbf{x}) \mathrm{d} m^0(\mathbf{x}')} = -[\mathcal{N}(\mathbf{x}|b_1, c_1), \ldots, \mathcal{N}(\mathbf{x}|b_M, c_M)] \beta \frac{\mathrm{d} \boldsymbol{\rho}}{\mathrm{d} m^0(\mathbf{x}')} \tag{21}$$

$$= -[\mathcal{N}(\mathbf{x}|b_1, c_1), \ldots, \mathcal{N}(\mathbf{x}|b_M, c_M)] \beta [\mathcal{N}(\mathbf{x}'|b_1, c_1), \ldots, \mathcal{N}(\mathbf{x}'|b_M, c_M)]^{\mathrm{T}} \tag{22}$$

Based on the above equation and (16), we have

$$V(\mathbf{x}, \mathbf{x}') = K(\mathbf{x}, \mathbf{x}') - \tilde{K}(\mathbf{x}, B) \beta \tilde{K}(B, \mathbf{x}') \tag{23}$$

Thus (13) holds. □

Blurring the Gaussian process by the local Gaussian distribution $\mathcal{N}(\mathbf{x}|b_i, c_i)$, we obtain the following corollary:

**Corollary 2** *The projection of the blurred Gaussian posterior process $q(f)$ onto $B$ is a Gaussian distribution with the following mean and covariance:*

$$\tilde{\mathbf{m}}_B = \hat{\mathbf{K}} \alpha \tag{24}$$

$$\tilde{\mathbf{V}}_B = \hat{\mathbf{K}} - \hat{\mathbf{K}} \beta \hat{\mathbf{K}}^{\mathrm{T}} \tag{25}$$

where $\hat{K}_{ij} = \int\int \mathcal{N}(\mathbf{x}|b_i, c_i) K(\mathbf{x}, \mathbf{x}') \mathcal{N}(\mathbf{x}'|b_j, c_j) \mathrm{d}\mathbf{x}\mathrm{d}\mathbf{x}'$.

Now the remaining question is how to estimate $(\mathbf{u}, \boldsymbol{\Lambda})$, or equivalently $(\alpha, \beta)$, such that the approximate posterior distribution on $B$ well approximates the exact posterior. In the following section, we describe an inference method for the sparse model.

### 3.2 Inference by expectation propagation

Expectation propagation (Minka, 2001) has three steps, message deletion, data projection, and message updates, iteratively applied to each training point. In the message deletion step, we compute the partial belief $q^{\backslash i}(f; m^{\backslash i}, v^{\backslash i})$ by removing a message $\tilde{t}_i$ (from the $i$-th point) from the approximate posterior $q^{\mathrm{old}}(f|m, v)$. In the data projection step, we minimize the KL divergence between $\tilde{p}(f) \propto p(t_i; f) q(f; m^{\backslash i}, v^{\backslash i})$ and the new approximate posterior $q^{\mathrm{old}}(f|m, v)$, such that the information from each data point is incorporated into the model. Finally, the message $\tilde{t}_i$ is updated based on the new and old posteriors.

Based on (5), the sparse GP is an exponential family with features $(\mathbf{g}_B(f), \mathbf{g}_B(f)\mathbf{g}_B(f)^{\mathrm{T}})$. As a result, we can determine the sparse GP that minimizes $KL(\tilde{p}(f)|q(f))$ by matching the moments on $\mathbf{g}_B(f)$.

Similar to (15) and (16), the moment matching equations are

$$\tilde{\mathbf{m}}_B = \tilde{\mathbf{m}}_B^{\backslash i} + \tilde{V}^{\backslash i}(B, \mathbf{x}_i)\frac{\mathrm{d}\log Z}{\mathrm{d}m^{\backslash i}(\mathbf{x}_i)} \tag{26}$$

$$\tilde{\mathbf{V}}_B = \tilde{\mathbf{V}}_B^{\backslash i} + \tilde{V}^{\backslash i}(B, \mathbf{x}_i)\frac{\mathrm{d}^2 \log Z}{(\mathrm{d}m^{\backslash i}(\mathbf{x}_i))^2}\tilde{V}^{\backslash i}(\mathbf{x}_i, B) \tag{27}$$

where $\tilde{V}^{\backslash i}(B, \mathbf{x})_j = \int \mathcal{N}(\mathbf{x}'|b_j, c_j)V^{\backslash i}(\mathbf{x}', \mathbf{x})\mathrm{d}\mathbf{x}'$ \hfill (28)

Combining (24) and (26) gives

$$\alpha = \alpha^{\backslash i} + \mathbf{b}\frac{\mathrm{d}\log Z}{\mathrm{d}m^{\backslash i}(\mathbf{x}_i)} \tag{29}$$

$$\mathbf{b} = \hat{\mathbf{K}}^{-1}\tilde{V}^{\backslash i}(B, \mathbf{x}_i) = \mathbf{p}_i - \beta_{\backslash i}\tilde{K}(B, x_i) \tag{30}$$

$$\mathbf{p}_i = \hat{\mathbf{K}}^{-1}\tilde{K}(B, x_i) \tag{31}$$

where we use (13) to obtain the last equation in the second line.

From (25) we have $\beta = \hat{\mathbf{K}}^{-1}(\hat{\mathbf{K}} - \tilde{\mathbf{V}}_B)(\hat{\mathbf{K}}^{-1})^{\mathrm{T}}$. Inserting (27) into this equation gives

$$\beta = \beta^{\backslash i} - \mathbf{b}\mathbf{b}^{\mathrm{T}}\frac{\mathrm{d}^2 \log Z}{(\mathrm{d}m^{\backslash i}(\mathbf{x}_i))^2} \tag{32}$$

These equations define the projection update. This update can be equivalently interpreted as multiplying $q^{\backslash i}(f)$ by an approximate factor $\tilde{t}_i(f)$ defined as:

$$\tilde{t}_i(f) = \mathcal{N}(\sum_j p_{ij}\int f(\mathbf{x})\mathcal{N}(\mathbf{x}|b_j, c_j)\mathrm{d}\mathbf{x}|g_i, \tau_i^{-1}) \tag{33}$$

$$\tau_i^{-1} = (-\nabla_m^2 \log Z)^{-1} - \tilde{K}(\mathbf{x}_i, B)\mathbf{b} \tag{34}$$

$$g_i = m^{\backslash i}(\mathbf{x}_i) + (-\nabla_m^2 \log Z)^{-1}\nabla_m \log Z \tag{35}$$

The approximation factor $\tilde{t}_i(f)$ can be viewed as a message from the $i$-th data point to the sparse GP. To check the validity of this update, we compute

$$\tilde{Z} = \int \tilde{t}_i(f)q^{\backslash i}(f)df \propto \mathcal{N}(u_i|\mathbf{p}_i^{\mathrm{T}}\tilde{\mathbf{m}}_B^{\backslash i}, \tau_i^{-1} + \mathbf{p}_i^{\mathrm{T}}\tilde{\mathbf{V}}_B^{\backslash i}\mathbf{p}_i) \tag{36}$$

$$= \mathcal{N}(u_i|m^{\backslash i}(\mathbf{x}_i), \tau_i^{-1} + \tilde{K}(\mathbf{x}_i, B)\mathbf{b}) \tag{37}$$

which has the same derivatives as the original $Z = \int t_i(f)q^{\backslash i}(f)df$. Therefore, the multiplication $\tilde{t}_i(f)q^{\backslash i}(f)$ leads to the same $q(f|m, v)$. In other words, we have $\tilde{t}_i(f) \propto q(f)/q^{\backslash i}(f)$.

To delete this message, we multiple its reciprocal with the current $q(f)$. Using the same trick as before, we can solve the multiplication using the following moment matching equations:

$$\mathbf{h} = \mathbf{p}_i - \beta\tilde{K}(B, \mathbf{x}_i) \tag{38}$$

$$\tilde{Z}_d = \int \frac{1}{\tilde{t}_i(f)}q^{\backslash i}(f)\mathrm{d}f \propto \mathcal{N}(u_i|\mathbf{p}_i^{\mathrm{T}}\tilde{\mathbf{m}}_B^{\backslash i}, -\tau_i^{-1} + \tilde{K}(\mathbf{x}_i, B)\mathbf{h}) \tag{39}$$

$$\frac{\mathrm{d}^2 \log \tilde{Z}_d}{(\mathrm{d}m(\mathbf{x}_i))^2} = -(-\tau_i^{-1} + \tilde{K}(\mathbf{x}_i, B)\mathbf{h})^{-1} \tag{40}$$

$$\frac{\mathrm{d}\log \tilde{Z}_d}{\mathrm{d}m(\mathbf{x}_i)} = (-\frac{\mathrm{d}^2 \log Z}{(\mathrm{d}m(\mathbf{x}_i))^2})(g_i - \tilde{K}(\mathbf{x}_i, B)\alpha) \tag{41}$$

$$\alpha^{\backslash i} = \alpha + \mathbf{h}\frac{\mathrm{d}\log \tilde{Z}_d}{\mathrm{d}m(\mathbf{x}_i)} \tag{42}$$

$$\beta^{\backslash i} = \beta - \mathbf{h}\mathbf{h}^{\mathrm{T}}\frac{\mathrm{d}^2 \log \tilde{Z}_d}{(\mathrm{d}m(\mathbf{x}_i))^2} \tag{43}$$

The inference algorithm is summarized in Algorithm 1.

---
**1.** Initialize $q(f)$, $g_i$, and $\tau_i$ all to be 0.
**2.** Loop until the change over all $g_i$, and $\tau_i$ is smaller
  than a threshold
   Loop over all training data point $\mathbf{x}_i$
    **Deletion.** Compute $\alpha^{\backslash i}$ and $\beta^{\backslash i}$ for $q^{\backslash i}(f)$ via (42)(43)
    **Projection.** Compute $\alpha$ and $\beta$ for $\mathbf{q}(f)$ via (29)(32)
    **Inclusion.** Update $g_i$, and $\tau_i$ for the message $\tilde{t}_i$ via (34)(35)

**Algorithm 1**: Expectation Propagation Inference for Sparse GP

---

### 3.3 Regression

Given the linear regression likelihood (2), the quantities in the projection step (29)(32) are

$$\frac{\mathrm{d}\log Z}{\mathrm{d}m(\mathbf{x}_i)} = \frac{y_i - m^{\backslash i}(x_i)}{v_y + v^{\backslash i}(x_i, x_i)} \qquad \frac{\mathrm{d}^2 \log Z}{(\mathrm{d}m(\mathbf{x}_i))^2} = \frac{-1}{v_y + v^{\backslash i}(x_i, x_i)} \tag{44}$$

### 3.4 Classification

Given the classification likelihood (3) where $\phi(\cdot)$ is the step function, the quantities in the projection step (29)(32) are

$$z = \frac{m^{\backslash i}(x_i) y_i}{\sqrt{v^{\backslash i}(x_i, x_i)}} \tag{45}$$

$$Z = \epsilon + (1 - 2\epsilon)\psi(z) \tag{46}$$

$$\frac{\mathrm{d}\log Z}{\mathrm{d}m(\mathbf{x}_i)} = \gamma y_i \tag{47}$$

$$\frac{\mathrm{d}^2 \log Z}{(\mathrm{d}m(\mathbf{x}_i))^2} = \frac{\gamma(m^{\backslash i}(x_i) y_i + v^{\backslash i}(x_i, x_i)\gamma)}{v^{\backslash i}(x_i, x_i)} \tag{48}$$

where $\gamma = \frac{(1-2\epsilon)\mathcal{N}(z|0,1)}{Z\sqrt{(v^{\backslash i}(\mathbf{x}_i))}}$ and $\psi(\cdot)$ is the standard Gaussian cumulative distribution function.

## 4 Related work

Most work in sparse GP approximation has been for regression with Gaussian noise, where the exact posterior distribution is a GP whose moments are available in closed form. The goal in that case is to approximate the exact GP with a simpler GP. Quinonero-Candela and Rasmussen (2005) compared several such approximations within a common framework, highlighting the FITC approximation as an improvement over DTC. Walder et al. (2008) later generalized FITC to include basis-dependent lengthscales, and showed that this approximation fits into the framework of Quinonero-Candela and Rasmussen (2005).

For classification with GPs, few papers have imposed an explicit sparsity constraint on the GP posterior. One reason for this is that sparse GP approximations have only been understood from a regression perspective. Csato and Opper (2000) showed how online GP classification could be made sparse, using an approximation equivalent to FITC. However, the FITC approximation was introduced as a subroutine, not presented as a part of EP itself. Csato (2002) later gave a batch EP algorithm, where a different sparse GP approximation (DTC) was used as a subroutine within EP. The software distributed by Csato can run either online or batch and has an option to use either FITC or DTC. In this paper, we clarify Csato and Opper's work by showing that the FITC approximation can in fact be viewed as part of the overall EP approximation. Furthermore we extend it to include VSGP approximation.

## 5 Experiments

We evaluate the new sparse GP method on both synthetic and real world data for regression and classification tasks and compare its predictive performance with alternative sparse GP methods. We use the RBF kernels for all the experiments. To all the sparse GP models, we need to choose basis centers. Instead of using the evidence framework, we use the entropy reduction criteria (equations are omitted to save space) to search over the hierarchical partitions of the data. We then use the cluster centers as the basis points and use the data partitions to compute the local covariance matrix. Our focus is on the approximate inference method so we use the same basis points for all the methods being evaluated.

We first examine the performance of the new method on synthetic data for the regression task. Since we can control the generative process of the synthetic data, it is easier for us to gain insight into how the method performs. For regression, we sample 100 data points along a circle with some additive Gaussian noise. The output in different quadrant has different values plus certain additive Gaussian noise.

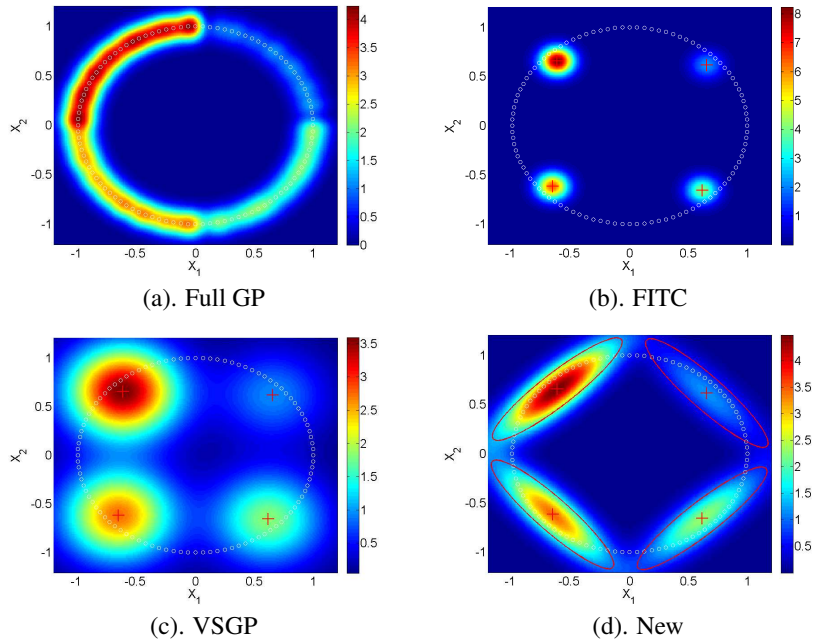

(a). Full GP  (b). FITC

(c). VSGP  (d). New

Figure 1: Illustration on simple circle data. The heatmaps represent the values of the posterior means of different methods. Red ellipses and crosses are the mean and the standard deviation of local covariances for the sparse GP model. The white dots are the training data points. EP estimation uses the full local covariance matrix in (d), significantly improving the estimation accuracy along the circle.

The mean of the exact posterior distribution of the full GP is shown in figure 1.a. The posterior means of FITC, SOGP, and the new method with sphere and full local covariance matrices. Four basis points are chosen as the centers learned by K-means clustering. Note that when the local covariance matrices become sphere matrices, our method reduces to the multiscale method (Walder et al., 2008). For FITC, we performed cross-validation to choose the right kernel width for the RBF kernel. As shown in the figure, the use of the full local covariance matrix improves the estimation accuracy.

Then we perform test FITC, SOGP, and the new methods with sphere and full local covariance matrices on a subset of the pumadyn-32nm dataset. On this dataset, however, we do not see the further improvement of our new method over FITC. This is consistent with the results reported in (Walder et al., 2008).

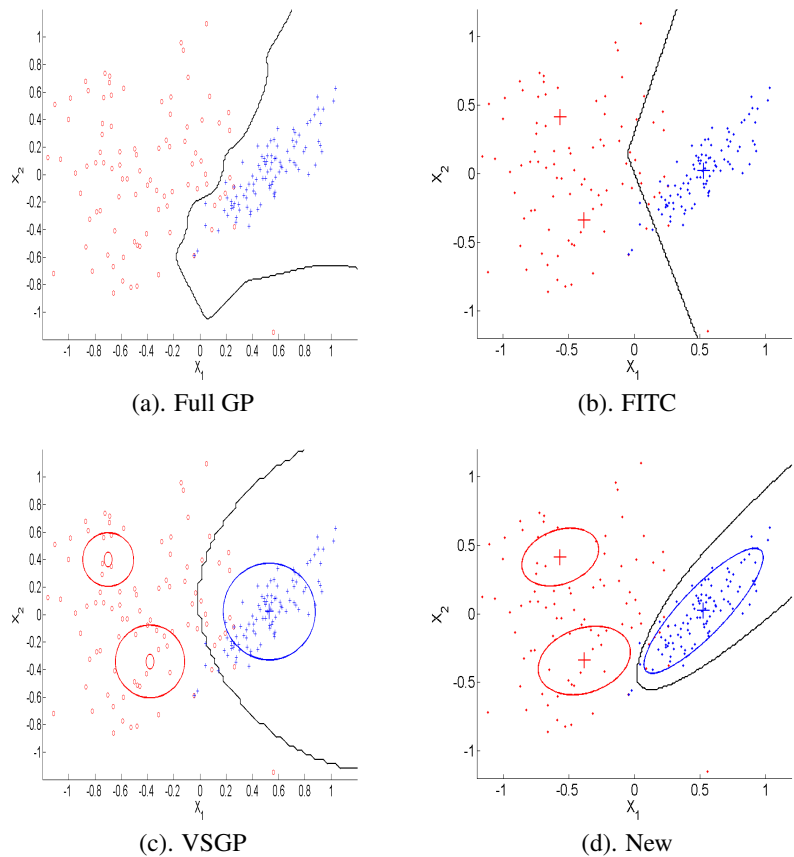

Figure 2: Classification on synthetic data. The blue and red ellipses show the standard deviation of local covariances for the sparse GP model. The black curve is the decision boundary. With only three basis points, the true, complex decision boundary in (a) is well approximated by an ellipse by our method (d).

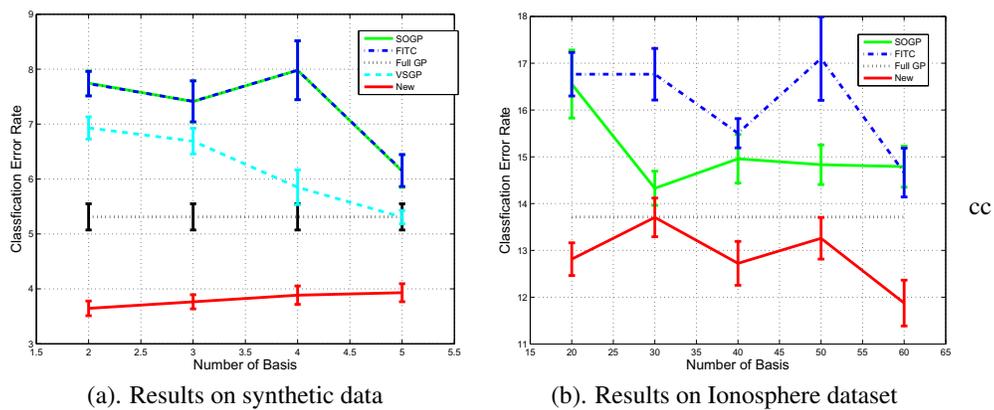

Figure 3: Classification on synthetic and real data.

We then test all the methods on synthetic data for a classification task. Each class of the data is sampled from a Gaussian and shown in 3. It also shows the decisions boundaries and the basis points. Figure 3 illustrates the advantages of our new approach. Since we simply use K-means to choose the basis points, further improvement for FITCis possible. We then sample 200 points for training and 2000 for testing. The experiments are repeated 10 times and the results are shown in figure 3. In this figure, we also report results of SOGP.

Finally we tested our methods on a standard UCI dataset, Ionosphere. The results are summarized in figure 3. Again, we outperform the previous methods.

## 6 Conclusions

In this paper, we present a new perspective to understand sparse GP models using the expectation propagation framework and develop new inference methods based on this perspective. Empirical results demonstrate improved prediction accuracy with the new extensions.